\documentclass[10pt, conference, letterpaper]{ieeeconf}   

\IEEEoverridecommandlockouts                              
\usepackage{subcaption} 
\usepackage{cite}
\usepackage{amsmath,amssymb,amsfonts}
\usepackage{graphicx}
\usepackage{textcomp}
\usepackage{caption}
\usepackage{float}
\usepackage{comment}
\usepackage{url}
\usepackage[table]{xcolor}
\usepackage{booktabs}
\usepackage{multirow}
\usepackage{colortbl}
\usepackage{listings}
\usepackage[ruled,vlined]{algorithm2e}
\usepackage{float}
\usepackage{placeins}

\lstdefinestyle{mystyle}{
    backgroundcolor=\color{white},   
    basicstyle=\footnotesize\ttfamily,       
    breakatwhitespace=false,         
    breaklines=true,                 
    captionpos=b,                    
    keepspaces=true,                 
    numbers=none,                    
    showspaces=false,                
    showstringspaces=false,
    showtabs=false,                  
    tabsize=4,
    frame=single,
    framesep=5pt,
    framexleftmargin=5pt,
    framexrightmargin=5pt,
    framextopmargin=5pt,
    framexbottommargin=5pt,
    xleftmargin=5pt,
    xrightmargin=5pt
}

\definecolor{ExcellentGreen}{RGB}{119,198,55}
\definecolor{GoodGreen}{RGB}{255,255,0}
\definecolor{PoorRed}{RGB}{255, 0, 0}
\definecolor{NoSignalRed}{RGB}{128, 0, 0}

\def\BibTeX{{\rm B\kern-.05em{\sc i\kern-.025em b}\kern-.08em
    T\kern-.1667em\lower.7ex\hbox{E}\kern-.125emX}}

\title{\LARGE \bf Meta-reasoning Using Attention Maps and Its Applications in Cloud Robotics}

\author{Adrian Lendinez$^{1}$, Renxi Qiu$^{1}$, Lanfranco Zanzi$^{2}$ and Dayou Li$^{1}$,
\thanks{$^{1}$Adrian Lendinez, Renxi Qiu and Dayou Li are with the University of Bedfordshire, LU1 3JU, University Square, Luton, UK {\tt\small Adrian.Lendinezibanez@study.beds.ac.uk, Renxi.Qiu@beds.ac.uk and Dayou.Li@beds.ac.uk}}%
\thanks{$^{2}$ Lanfranco Zanzi is with NEC Laboratories Europe, Heidelberg, DE {\tt\small lanfranco.zanzi@neclab.eu}}
}

\begin{document}
\maketitle

\begin{abstract} 
Meta-reasoning, a branch of AI, focuses on reasoning about reasons. It has the potential to enhance robots' decision-making processes in unexpected situations. However, the concept has largely been confined to theoretical discussions and case-by-case investigations, lacking general and practical solutions when the Value of Computation (VoC) is undefined, which is common in unexpected situations. In this work, we propose a revised meta-reasoning framework that significantly improves the scalability of the original approach in unexpected situations. This is achieved by incorporating semantic attention maps and unsupervised "attention" updates into the meta-reasoning processes. To accommodate environmental dynamics, "lines of thought" are used to bridge context-specific objects with abstracted attentions, while meta-information is monitored and controlled at the meta-level for effective reasoning. The practicality of the proposed approach is demonstrated through cloud robots deployed in real-world scenarios, showing improved performance and robustness.
\end{abstract} 

\section{Introduction}
\label{sec:intro}

Significant progress has been made in probabilistic robotics to improve the adaptability and robustness of robot operations~\cite{Thrun2005}. By integrating probabilistic models and statistical methods into perception and decision-making processes, robots can address structured uncertainty and randomness. However, to remain robust in unexpected situations, autonomous systems must also manage their reasoning processes, such as effectively handling uncertainties at the ground level and adapting objects at the conceptual level. This capability, known as meta-reasoning, facilitates reasoning about reasons~\cite{Herrmann2023}. Although meta-reasoning has been a topic of discussion in AI for many years~\cite{Stuart1990,Cox2013,Griffiths2019}, it has not been fully integrated into ad hoc robotic operations due to the lack of scalable computational models~\cite{Herrmann2023,Conitzer2013} in unstructured or unexpected situations. To address this challenge, this paper introduces the concept of a semantic attention map to enhance traditional meta-reasoning mechanisms. Using attention-based unsupervised belief updates, the proposed approach attempts to scale the computational model of meta-reasoning when conceptual objects are unstructured. The improved scalability of meta-reasoning will enable robots to reason more effectively in unexpected situations.

Computational models for meta-reasoning address the exploration vs. exploitation dilemma by allowing robots to assess whether the benefits of maintaining current beliefs \( b \) outweigh the time and resources required to expand their knowledge base \( b' \). This process, which involves calculating the expected utility function \( \mathbb{E}[U(a) \mid b] \), is captured by the Value of Computation (VoC), formally defined in~\cite{Griffiths2019}. 

One of the main impediments of practical meta-reasoning has been the symbolic grounding problem~\cite{Qiu2012,Liu2012}, such as the difficulty in quantifying beliefs into concrete symbols to calculate the Value of Computation (VoC) and the operational costs, especially in unexpected situations where context is not predefined. To address this issue, we propose Bayesian-based semantic attention maps to manage hypotheses through attention in an unsupervised fashion. This innovation bridges the gap between ground-level reasoning and meta-level reasoning via lines of thought to map rewards loosely into the attention, thereby avoiding the need for detailed context on symbolic grounding. This provides a scalable and efficient protocol for closed-loop meta-reasoning in partially defined territories by tracking domain-specific symbols and context-independent attention through quality experience replay.

As a specific example of proactive robot reasoning, cloud robotics provides a context in which attention-enhanced meta-reasoning is applied and verified. In cloud robotics, various factors such as network signal quality (e.g., RSRP, RSRQ metrics) and edge handover introduce unexpected situations and out-of-distribution observations. Our framework integrates these unforeseen variables into the decision-making loop, enabling informed and scalable decisions. The improved scalability and practicality of meta-reasoning will be demonstrated in real-world robotic systems, measured by enhanced robustness and quality of experience for cloud robotics.

\vspace*{0.5mm}

This paper is structured as follows.
Sec.~\ref{sec:related} summarizes the related works in the field.
Sec.~\ref{sec:problem} outlines the primary challenges of implementing meta-reasoning in practical robotic applications, particularly scaling the reasoning to handle unexpected situations.
Sec.~\ref{sec:solution} describes the proposed solution.
Sec.~\ref{sec:eval} demonstrates the proposed solution in cloud robotics scenarios and evaluates its potential benefits for robustness and performance metrics in unexpected situations.
Finally, Sec.~\ref{sec:Conclusion}
concludes this paper.

\section{Related Works} 
\label{sec:related}

 Meta-reasoning offers meta-level control of computational activities and introspective monitoring of reasoning~\cite{Herrmann2023}. In addition to reasoning about the ground truth, it helps robots reason about emerging concepts and switch among different control policies. The integration of meta-reasoning into decision-making processes is a growing area of research in both human cognition and artificial intelligence. While traditional decision-making frameworks in robotics often rely on fixed rules or algorithms, meta-reasoning allows systems to adapt their reasoning strategies based on context, prior experiences, and available resources. This approach, which can be seen as "thinking about thinking," has been widely discussed in cognitive science and machine learning. In~\cite{Conitzer2013}, the authors emphasize the importance of adaptive strategies in decision-making, where an agent assesses not only what actions to perform, but also how to allocate cognitive resources (e.g., time, effort) to the reasoning process itself. Both humans and machines benefit from prior knowledge and past experiences to refine their decision-making processes. This enables more efficient exploration of problem spaces and avoidance of suboptimal solutions - an issue that is particularly relevant in robotic systems operating in dynamic, resource-constrained environments. This concept has been successfully applied in neuroscience through meta-analysis, which synthesizes quantitative data from multiple independent models~\cite{Gurevitch2018}.

In~\cite{Epstein2013} the authors identified variants of meta-reasoning. Suppose there is a fixed set of possible plans of action (decisions) that the agent can choose from, each with an expected utility. The agent can take deliberation actions on each plan, and depending on the rewards of these deliberation actions, the expected utility of the plan changes. The goal is to find a deliberation strategy that maximizes the agent's expected utility/reward. Another variant involves the agent knowing that there are multiple states in clarification/prediction. To obtain nonzero utility, the agent must determine the state with certainty through deliberation. Unlike meta-learning~\cite{Griffiths2019,Kumar2024}, which focuses on the efficient use of data at the ground level for modeling, meta-reasoning emphasizes the efficient use of computational models and resources at the object level~\cite{Epstein2013}. This enables meta-level explainability of objects and concepts in unstructured environments~\cite{Cox2013}. Therefore, the concept of meta-reasoning, or "reasoning about reasoning," aligns to enhance robust robot planning in unexpected situations. By projecting the experience of robots into the reward space, meta-level explainability is enabled via reward-driven optimisation independent of ground operations.

Although meta-reasoning provides a foundational understanding of optimizing decision-making by decoupling the adaptation of ground truth from the evolution of symbolic concepts, its practical application in robotics remains limited. The latest effort can be found in ~\cite{Parashar2018}, where meta-reasoning is introduced into lifelong learning through Hierarchical Task Network (HTN). The meta-reasoner identifies mismatches between the current task and the original task, generating goals and rewards to reduce the mismatch. Additionally, ~\cite{Herrmann2023} focuses on integrating the meta-reasoning architecture into robot operations on a case-by-case basis by manually selecting the reasoning processes that the meta-level should control. Despite success in specific applications, these solutions are not yet fully scalable. This limitation arises from the conflict between context-specific reasoning at the ground level and the need for seamless integration into context-independent generalisation in unexpected situations. An unsupervised knowledge generalisation approach is urgently needed for traditional meta-reasoning to be applied to the intended use cases. 

\section{Problem Statement}
\label{sec:problem}
To extend robot reasoning from structured and controlled laboratory environments to unstructured and unpredictable real-world scenarios, meta-reasoners must be scalable at the architectural level for practical robotic operations. A key challenge lies in balancing the trade-off between context-aware reasoning and context-independent reasoning:
\textbf{Context-Aware Reasoning:} For effective decision-making, robotic reasoning must be context-aware and tightly coupled with ground-level policies. The Value of Context (VoC), represented through symbols and rules, must be accurately grounded in real-world actions and sensory perceptions to ensure practical applicability.\\
\textbf{Context-Independent Reasoning:} At the same time, robotic reasoning must maintain context independence at the symbolic level, allowing for the adaptation of VoC based on newly observed phenomena. This generalization process should be executed in an unsupervised manner to accommodate unforeseen situations.
In this context, a fundamental challenge arises: if the type of context-independent generalization were already anticipated and well-defined at the symbolic level, the situation would no longer be considered "unexpected." This creates a classic chicken-or-egg dilemma that is difficult to resolve. However, within the meta-reasoning framework, context-aware reasoning is decoupled from context-independent reasoning. The former corresponds to object-level reasoning for executing ground actions, while the latter pertains to meta-level reasoning for reasoning about decisions, as illustrated in Fig.~\ref{fig:traditional}.
In addition to the context dilemma, meta-reasoning in robotics faces a significant observability challenge. Meta-reasoning governs policies at the object level, such as offloading decisions and edge-switching in cloud robotics. Meanwhile, object-level reasoning governs policies at the ground level, including navigation and object detection. 
Crucially, neither the rewards associated with meta-level decisions nor those linked to object-level decisions are directly observable. Instead, only the overall reward, which emerges as a consequence of the interplay between meta-level and object-level decisions, can be directly measured. This interdependency necessitates sophisticated techniques for credit assignment and learning in real-world robotic applications.

Under these circumstances, the meta-reasoning problem for robots can be defined as follows:
\begin{itemize}
    \item \textbf{Q1} How can meta-reasoning frameworks bridge the gap between ground-level, context-aware operations and meta-level, context-independent VoC generalization?
    \item \textbf{Q2} How can scalable meta-reasoning architectures enable effective unsupervised computational models, allowing meta-decisions to adapt dynamically without requiring manual definition of reasoning objects on a case-by-case basis?
\end{itemize}

\section{Proposed solution}
\label{sec:solution}
\begin{figure}[t]
    \centering
    \begin{subfigure}[b]{0.9\columnwidth} 
        \centering
        \includegraphics[width=0.9\linewidth,clip,trim=0cm 0cm 0cm 0cm]{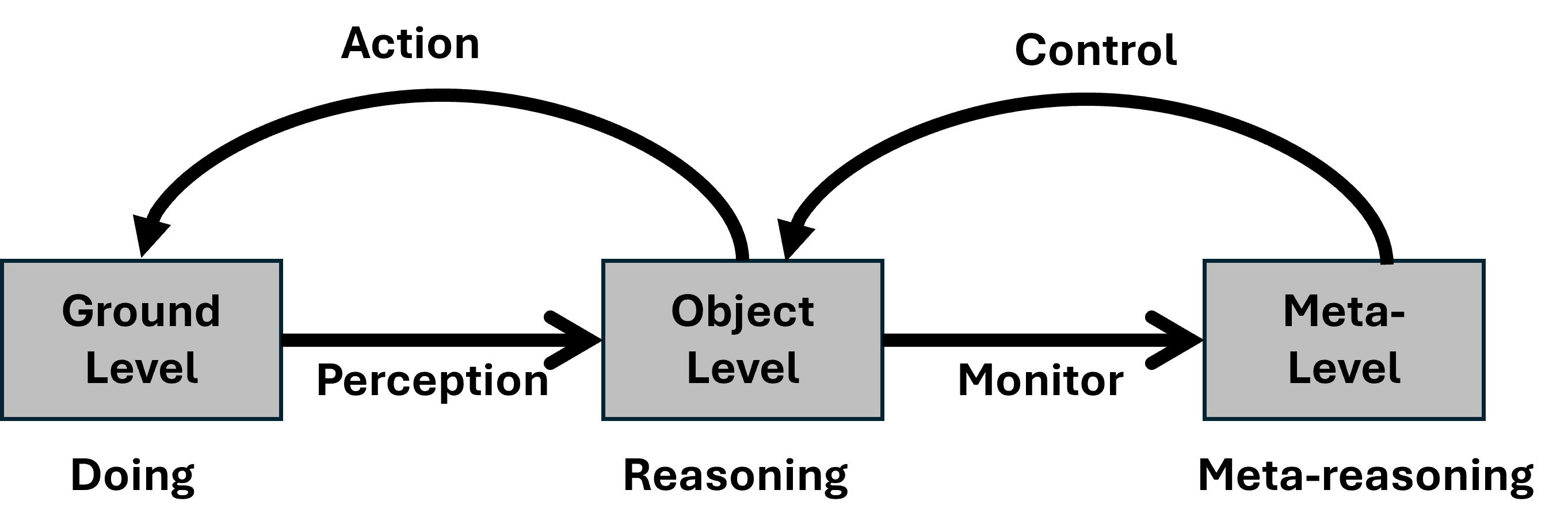}
        \caption{\textit{The traditional meta-reasoning from Herrmann~\cite{Herrmann2023}}}
        \label{fig:traditional}
        \vspace{2.2mm}
    \end{subfigure}
    \vspace{2mm}
    \begin{subfigure}[b]{0.9\columnwidth} 
        \centering
        \includegraphics[width=0.9\linewidth,clip,trim=0cm 0cm 0cm 0cm]{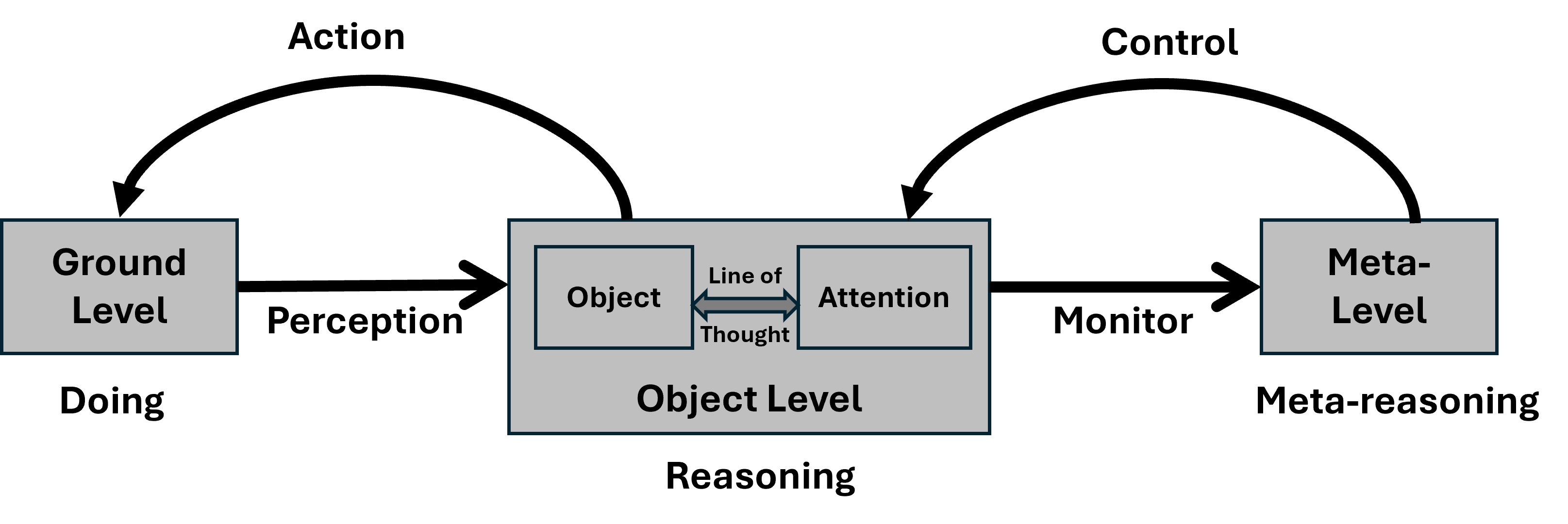}
        \caption{\textit{The proposed meta-reasoning computational model with attention}}
        \label{fig:proposed}
    \end{subfigure}
    \vspace{0.5mm}
    \caption{Meta-reasoning computational models.}
    \label{fig:scenario2}
\end{figure}

To address these limitations, our approach extends the traditional meta-reasoning framework by simplifying object-level conceptualization through a context-independent symbol called \textit{attention}. The original meta-reasoning computational model, illustrated in Fig.~\ref{fig:traditional}, has been modified by decoupling the fully coupled meta-reasoning and object-reasoning loops into two loosely connected yet independent reasoning loops, as shown in Fig.~\ref{fig:proposed}. 
In this revised model, the meta-reasoning loop (right-hand side of Fig.~\ref{fig:proposed}) focuses on context-independent \textit{attentions}, while the object-reasoning loop (left-hand side) handles context-specific ground-level information. These loops communicate through a concept known as the \textit{line of thought}, which facilitates the interaction between meta-level and object-level decisions in deriving the final Value of Control (VoC). 
Attention is a probabilistic metric, characterized by a mean and standard deviation, designed to estimate the potential impact or rewards of decisions. It is further categorized into:
 \emph{i)} the \textbf{meta-attention} which measures the impact/rewards of meta-level decisions. It is context- and object-independent, derived indirectly from the experienced VoC and filtered by the line of thought. Meta-attention can be updated using unsupervised algorithms.
\emph{ii)} the \textbf{ground-attention} which measures the impact/rewards of robot capabilities. It is context- and object-specific, also derived indirectly from the VoC and filtered by the line of thought.
Notably, since meta-attention is independent of specific objects, it avoids the traditional chicken-or-egg dilemma of requiring pre-defined, hard-coded symbols for unexpected situations, thereby enabling true adaptive reasoning.\\
Nevertheless, a key challenge in adaptive reasoning is generating VoC when the contextual symbols that define a robot’s control are unknown or ambiguous. While object-reasoning typically requires context-specific information, meta-reasoning remains flexible by replacing traditional symbolic grounding with the more adaptable \textit{line of thought}. This decomposition allows solutions to be dynamically adjusted within the object context, enhancing flexibility in task resolution and improving the robot's ability to adapt to unforeseen scenarios.

\subsection{Semantic Attention Map}
To fully integrate the proposed meta-reasoning into ad hoc robot operations, attention within the proposed meta-reasoning framework is implemented using attention maps. These maps are a novel type of semantic map that adds an extra layer to direct the robots' focus toward specific areas while quantifying beliefs or hypotheses about them. Within this framework, these maps control symbolic attention during robot operations, converting grounded rewards into records of the robot’s experience. Formally, for each cell in the grid, attention $A$ is initialized over the symbol $S$, and then represented as a belief distribution:
\begin{equation}
    P(A | D) = \frac{P(D | A) P(A)}{P(D)},
\end{equation}
where $P(A|D)$ represents the updated belief about the utility $A$ given the historically observed utility $D$, and $P(D|A)$ indicates the likelihood of the data under the current object associated with the line of thought.

Unlike traditional occupancy grid maps, which store only data about physical obstacles, semantic maps extend this capability by representing abstract concepts. Attention maps further enhance this by assigning simple high (green) or low (yellow) attention values, directing the robot’s focus toward relevant areas. These attention maps are generated using the \texttt{pointcloud2} ROS2 data structure, as described in \cite{5GERA_IROS,5GERA2024}. A key advantage of this approach is that temporal beliefs can be converted into attention values without being explicitly grounded in predefined contextual objects. Attention values can be updated in an unsupervised manner using a Bayesian framework. The attention map acts as an interface between the robot's perceived experience and the dynamics of the environment, facilitating the integration between the meta-reasoning loop and robot experiences using Bayesian inference.
Specifically, a beta update mechanism introduced in the next section, enables the calculation of high or low attention ratios based on the robot’s observations and the underlying hypothesis.


\subsection{Unsupervised update}
The attention map is dynamically updated using a Bayesian approach, where regions of high and low attention follow a binomial distribution. This distribution is refined through a Beta distribution update, ultimately transforming beliefs into a mean ($\mu$) and variance ($\sigma$) for subsequent reasoning. 
For example, if an area is initially assigned high attention based on a hypothesis (e.g., prioritizing signal strength), this assumption can later be adjusted to low attention if empirical observations indicate that signal quality has minimal impact on task performance. This adaptation is driven by a utility function that integrates cost and reward components, optimizing the decision-making process.
%
As the robot operates, it dynamically updates the attention associated to symbols using a Beta distribution:
\begin{equation}
    P_{\text{attention}}(A) \sim \text{Beta}(\alpha, \beta)
\end{equation}
where parameters $(\alpha)$ and $(\beta)$ evolve based on success or failure in tasks, while adjusting the attention levels accordingly.
The pseudo-code for the revised meta-reasoning using attention is presented in Algorithm~\ref{algo:attention_map}.

\begin{footnotesize}
\begin{algorithm}[!t]
\footnotesize
\caption{Attention-Based Meta-Reasoning}
\label{algo:attention_map}

\SetAlgoLined

\textbf{STEP 1: Convert Existing Line of Thoughts into a Set of Rules} \\
Define \texttt{line\_of\_thought} as an empty set \\
\ForEach{\texttt{RULE\_X} in \{\texttt{RULE\_1, RULE\_2, ..., RULE\_N}\}}{
    Store \texttt{RULE\_X} in \texttt{logical\_structure}
}

\textbf{STEP 2: Define General Attention Categories} \\
Define \texttt{ground\_objects} = \{\texttt{OBJECT\_1, OBJECT\_2, ..., OBJECT\_M}\} \\
Define \texttt{meta\_objects} = \{\texttt{META\_1, META\_2, ..., META\_K}\} \\
\textbf{STEP 3: Extract Origin and Destination of the Attention from STEP2 for Attention Updates} \\
Define thought = line\_of\_thought
Define \texttt{origin} = extract\_origin\_objects(\texttt{thought})  // Extract preconditions \\
Define \texttt{destination}= extract\_destination\_objects(\texttt{thought})  

\textbf{STEP 4: Initialize Probabilistic Bayesian Parameters and Attention Map with Logical Consistency} \\
\ForEach{\texttt{grid} in \{\texttt{map}\}}{
    Set $\mu_0$ = \texttt{INITIAL\_MEAN}, $\sigma_0$ = \texttt{INITIAL\_VARIANCE} \\
}
\ForEach{\texttt{obj} in \{\texttt{origin} $\cup$ \texttt{destination}\}}{
    Set \texttt{attention\_map[obj]} with initial values \\
    Adjust based on dependencies in \texttt{line\_of\_thought}:
    \begin{itemize}
        \item If dependencies have LOW attention, decrease $\mu$.
        \item If dependencies have HIGH attention, increase $\mu$.
    \end{itemize}
}

\textbf{STEP 5: Iterative Attention Map Update} \\
\texttt{counter\_episode} = 1

\While{\texttt{task\_success} $<$ \texttt{success\_threshold}}{
    
    //\textit{Compute impact based on thought chain execution and reward. Ensure attention is properly placed over time and over the meta parameters.} 

    counter\_episode = counter\_episode + 1

        // No convergence over time
        
    \texttt{\textbf{IF}} (counter\_episode == 50) AND 
         \texttt{\textbf{IF}}({check(convergence) == \textbf{False}})   \texttt{\textbf{THEN}}
             update(impact\_weight), counter\_episode = 0

    \texttt{impact} = Impact\_calculation(\texttt{line\_of\_thought}, status of \texttt{origin}, impact\_weight ) \\
    \ForEach{\texttt{obj} in \{\texttt{origin} $\cup$ \texttt{destination}\}}{
        $\mu_{new}, \sigma_{new}$ = gaussian\_update(\\ \texttt{attention\_map[obj].$\mu$},\\ \texttt{attention\_map[obj].confidence}, \texttt{impact})
    }
    counter\_episode = counter\_episode +1
}
\textbf{STEP 6: Return Attention Map and Revised Impact\_Weight} \\
\Return \texttt{Updated attention\_map, impact\_weight}
\end{algorithm}
\end{footnotesize}
%
This code is designed to update attention in an unsupervised manner based on a predefined reward function while attempting to ground the updated attention within the preexisting line of thought objects if possible. The latter process requires identifying the appropriate attention-objects that reflect trends in the reward function. 
The initial attention distribution should match the distribution of the line of thought. However, due to system dynamics, attention distribution must be gradually updated over time. In certain cases, the logarithm can retain the mapping but may fail to establish a meaningful grounding with the objects in the line of thought. If attention is not grounded in any object, reasoning can still proceed, albeit with suboptimal explanability.

\subsection{Revised meta-reasoning computational model}


As shown in Fig.\ref{fig:traditional}, meta-reasoning model consists of two primary components: 1) Control and monitoring of objects, which tracks the effectiveness of symbolic representations in guiding behavior, and 2) VoC for meta-level strategy alternation, which quantifies new benefits into object-level symbols to reflect the latest environmental changes. However, in unexpected situations, the association between newly established abstractions and existing symbolic objects may not always be available, subsequently breaking the iteration of the meta-reasoning loop for VoC calculation defined in the literature~\cite{Griffiths2019} as:

\begin{equation}
\begin{aligned}
\text{\textit{VoC}}(c, b) = \mathbb{E}_{p(b' \mid b, c)} \left[
\max_{a'} \mathbb{E}\bigl[ U(a') \mid b' \bigr] 
- \max_{a} \mathbb{E}\bigl[ U(a) \mid b \bigr] \right]
- \text{cost}(c)
\end{aligned}
\end{equation}

where $a$  represents the robot's action, $b$ denotes the belief, and $c$ is the cost of operation over the belief. In this scenario, if there is no grounding for the updated belief $b'$ the expected new utility $\mathbb{E} \left[ U(a') \mid b' \right]$ cannot be calculated, deeming the equation unsolvable. Instead of waiting for the belief to be grounded to a specific object, which is not necessarily feasible all the time, it is worthwhile to hold a temporal abstraction called attention to store the dynamics of the reward and then gradually translate the attention back to control objects after sufficient confidence is established. Therefore, the traditional meta-reasoning loop is divided into two separate loops as described in the beginning of the section.\\
The Value of Computation (VoC) determines whether re-evaluating an attention strategy provides a performance benefit. It quantifies the expected improvement in task efficiency by shifting attention from one symbolic hypothesis to another.
Given a robot performing object detection, VoC is calculated as:
\begin{equation}
    VoC = U_{\text{best new}} - U_{\text{current}} - cost
\end{equation}
where \( U_{\text{current}} \) is the expected utility of the current attention strategy, and \( U_{\text{best new}} \) is the expected utility of the best alternative strategy.

To better trace the dynamics of the abstraction, the individual impact of the newly established attention needs to be extracted from the observed overall VoC. This can be done using pre-defined decomposition between ground-level capabilities and meta-level attentions, as illustrated in Algorithm~\ref{algo:attention_map}.  The capability of the robot is normally well-defined with significantly fewer unexpected situations compared to environmental changes. As shown in the evaluation section, this assumption is consistent with our experimental results. Using Retrieval Augmented Generation (RAG) with LLM~\cite{RAG2020} or dynamic fuzzy/logical equations, automated solutions can also be developed in the future to further enhance scalability.

\section{Applications and evaluation}
\label{sec:eval}
In this section, various cloud robotics scenarios are selected to evaluate the performance and robustness of the proposed attention-based meta-reasoning framework. The behavior of cloud robots in these scenarios is influenced by both ground-level control policies—such as navigation and perception—and meta-level policies that govern task allocation between local/onboard operations and cloud-based remote processing. Additionally, these policies facilitate coordination with nearby edge nodes to optimize resource utilization.
While ground operations are well-defined based on the robots' inherent capabilities, the surrounding environment remains highly unpredictable. This variability makes these scenarios particularly well-suited for benchmarking planning algorithms in unexpected situations, providing a rigorous evaluation of the adaptability and efficiency of the proposed meta-reasoning approach.
Three meta-reasoners were designed for the test, each representing a typical meta-reasoner case:
\begin{itemize}
    \item \textbf{R1}: Generic meta-reasoning~\cite{Cox2013} with no case-specific meta-knowledge. The VoC is defined based on location, which is extracted from historical experience of expected situations: VoC = f(Location(X)).
    \item \textbf{R2}: Customized meta-reasoner with case-specific knowledge on radio network quality. The VoC function is manually adjusted as described by Herrmann ~\cite{Herrmann2023}, where VoC = f(Signal-Quality(X)).
    \item \textbf{R3}: The proposed meta-reasoner is enabled by a semantic attention map, with VoC = f(Attention(X)).
\end{itemize}
The performance of the three reasoners is measured in two case studies using the following KPIs:
\begin{itemize}
    \item \textbf{KPI1: Success Rate (\%)}:  A task is considered successful if the robot fulfills the given task correctly.
    
    \item \textbf {KPI2: Battery Consumption(\%)}: The corresponding battery life consumption per given task completion, for example, per uniquely detected object.
    
    \item \textbf{KPI3: Robustness}: The system's ability to maintain stable performance under varying conditions and understand the logic behind successful performance, thereby processing data properly.

    \item \textbf{KPI4: Availability(\%)}: The proportion of time during which the service is available relative to the total operational time.
\end{itemize}

\begin{figure}[h]
    \centering
    \subfloat[Robotnik's Summit XL]{%
        \includegraphics[width=0.45\linewidth]{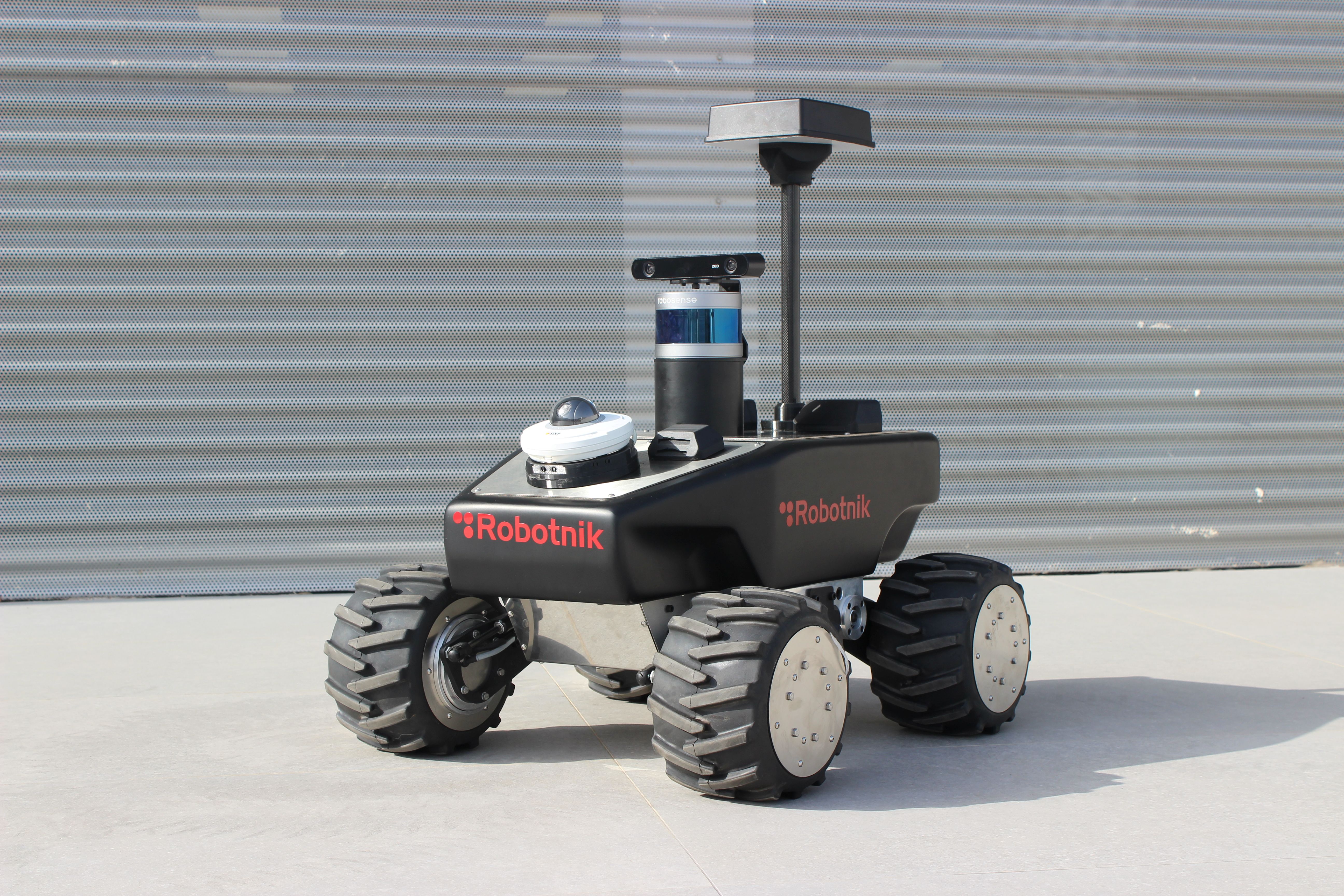}%
        \label{fig:robotnik}
    }
    \hfill
    \subfloat[Amarisoft testbed]{%
        \includegraphics[width=0.45\linewidth]{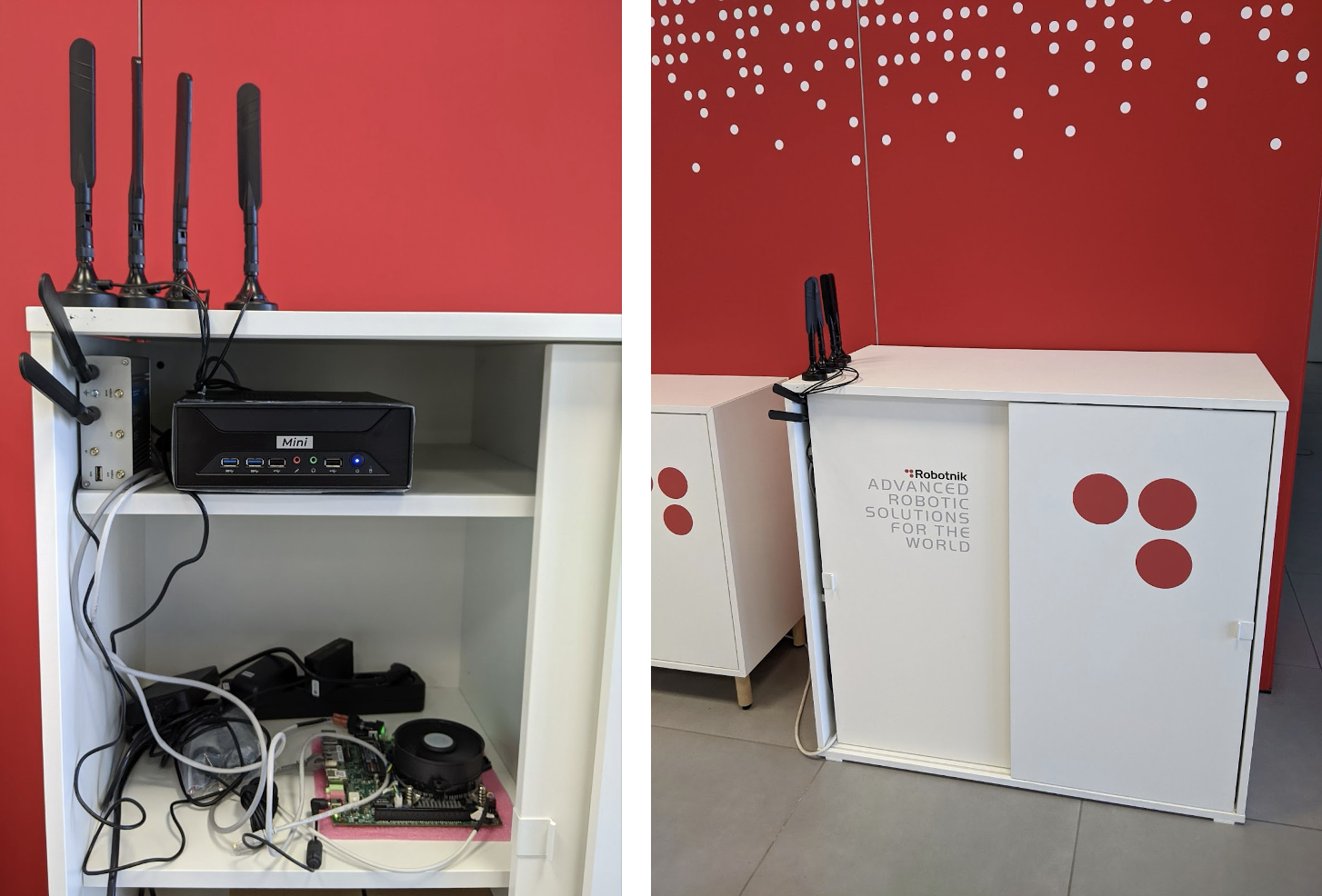}%
        \label{fig:testbed}
    }
    \vspace*{3.5mm}
    \hfill
    \subfloat[Signal quality map generation]{%
        \includegraphics[width=0.45\linewidth]{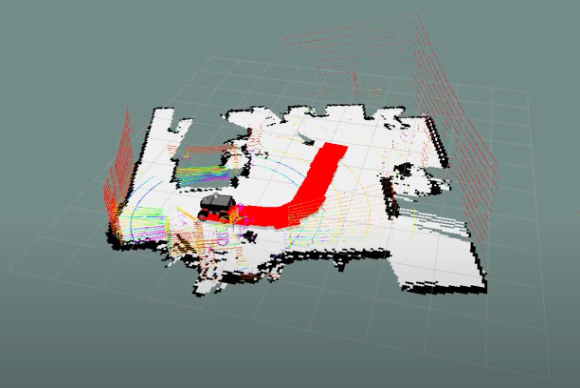}%
        \label{fig:signal_map}
    }
    \hfill
    \subfloat[Semantic map of the testing facility]{%
        \includegraphics[width=0.45\linewidth]{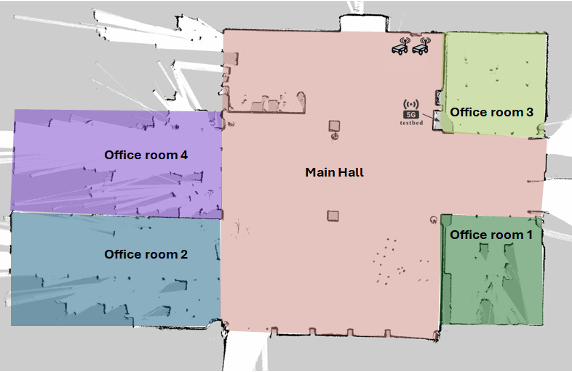}%
        \label{fig:semantic_map}
    }
    
    \vspace*{2.0mm}
    
    \caption{Experimental setup: (a) Robotnik's Summit XL, (b) Amarisoft testbed, (c) Signal quality map generation, and (d) Semantic map of facilities for experiments.}
    
    \label{fig:experimental_setup}
    
\end{figure}

As illustrated in Fig.\ref{fig:robotnik}, the robot used was a Robotnik's Summit XL~\cite{SUMMIT-XL} equipped with ROS 2 Humble and a 5G and WiFi antenna. To measure signal quality, RSRP, RSSI, and RSRQ radio metrics were used. The gNB (5G base station) used for testing was an Amari Callbox mini~\cite{Amarisoft}. Fig.\ref{fig:testbed}, Fig.\ref{fig:signal_map} and Fig.\ref{fig:semantic_map} depict the testing facilities with its corresponding semantic map. To perform the experiment, the Connected Robotics Platform (CRoP) \cite{ROSWiki} is used to dynamically offload onboard services such as the object detection pipeline to nearby edge devices, or towards remote cloud. 
Additionally, the CRoP platform is utilized to perform fully orchestrated handshakes and switch-overs. The attention maps are generated using the \texttt{pointcloud2} ROS2 data structure, following the methodology presented in~\cite{5GERA_IROS, SemanticMapROSPacakge}. These attention maps employ the same technology as signal quality maps, which represent another type of semantic map that seamlessly integrates with ROS.

\subsection{\textbf{Case Study 1: Mobile Robots Off-loading for Enhanced Object Detection}}

\begin{figure}[h]
\centering
\includegraphics[width=.9\columnwidth,clip, trim = 0cm 0cm 0cm 0cm]{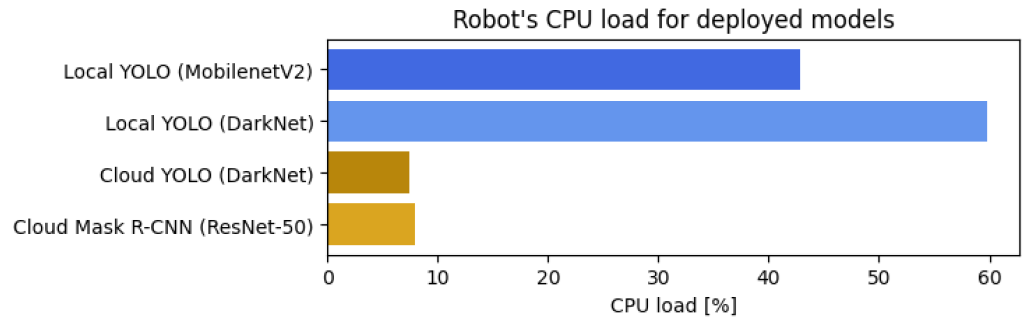}
\caption{CPU consumption among local/cloud processing}
\label{fig:overview}
\vspace{2mm} 
\end{figure}

Experimental results from \cite{5GERA2024} demonstrate that offloaded object detection achieves a higher success rate and lower battery consumption compared to local resource execution. In this case study, the robot's task is to detect and recognize various objects, with the key \textbf{KPI1} being being specifically defined as the number of successfully detected unique objects per task. There are 50 possible unique objects, randomly distributed across rooms. YOLO (DarkNet) is applied on edge machines for cloud-based object detection, enabling high success rates and low resource consumption. Meanwhile, on-board object detection is available using YOLO (MobilenetV2) in case the network connection is unstable. The difference in CPU workload is presented in Fig.\ref{fig:overview} along with the corresponding trends in battery consumption as of \textbf{KPI2}. In this experiment, the DarkNet detector configured in the edge/cloud can handle all 50 possible objects, while the MobilenetV2 detector can only handle a subset of these objects. The remaining objects are considered out-of-distribution for the on-board detector.

In case the robot gets stuck in a local minimum, help from an external human operator will be applied with the following interventions:
\begin{itemize}
    \item \textbf{Intervention Type 1}: Switch the offloading policy.
    \item \textbf{Intervention Type 2}: Manually drive the robot to another area.
\end{itemize}

A robust task planner would require very little intervention by the human operator. If the robot can recognize and utilize the additional information provided by human intervention, it is also considered to be achieving robustness to some extent. Under this presumption, \textbf{KPI3} (Robustness) is quantified using the equation below:
\begin{equation}
 Robustness = 100 - RI - UI * 3
\end{equation}
where \textbf{\textit{RI}} represents the number of Recognizable Interventions and \textbf{\textit{UI}} represents the number of Unrecognised Interventions.
%
Table~\ref{tab:comparison} summarizes typical unexpected situations that cause failures in this case study. Additionally, the expected capabilities of the reasoners (R1-R3) are listed in the same table. The actual performance of the reasoners will be benchmarked against KPI1-KPI3 to determine if they align with the anticipated capabilities. Within the table:
\begin{itemize}
    \item \textbf{Yes}: reasoner should be capable of addressing the situation. For example, adjusting the meta-level control policy to mitigate the unexpected situation.
    \item \textbf{Partial}: reasoner could partially address the unexpected situation, but limited to exploring solutions within pre-defined context.
    \item \textbf{No}: reasoner cannot handle the unexpected situation.
\end{itemize}

\begin{table}[t]
    \caption{\small List of potential failure causes and reasoning processes to address the failure.  \textbf{(R1)} - traditional meta-reasoning, \textbf{(R2)} - traditional meta-reasoning with manually defined signal quality object and \textbf{(R3)} -our proposed meta-reasoning with attention map.}
    \centering
    {%
    \begin{tabular}{  m{4.5cm}  m{0.8cm}  m{0.8cm}  m{0.8cm} } 
        \hline
        \toprule
        \textbf{Unexpected situations} & \textbf{R1} & \textbf{R2} & \textbf{R3}  \\
        \midrule
        Same objects detected repetitively & Partial & Yes & Yes \\
        Poor signal quality & No & Yes & Yes \\
        Edge computing overload  & Partial  & Yes & Yes \\
        Out of distribution objects  & Partial  & Yes & Yes \\
        Poor detection due to robot location  & Partial & Partial & Yes \\
        \midrule
    \end{tabular}%
    }
    \label{tab:comparison}
\end{table}

Initially, the robot is provided with the knowledge that the success rate of cloud-based object detection is highest in the Main Hall. Therefore, VoC = f(Location(Main Hall)) is expected to be higher than in other areas. This is reflected in the initial belief over the line of thought:

\begin{small} 
\begin{itemize}
    \item \textbf{IF} (NAVIGATION AND DETECTION) \textbf{THEN} MAIN\_HALL: \textit{The robot requires both a working navigation and detection system to establish that it is in the main hall}.
    \item \textbf{IF} MAIN\_HALL \textbf{THEN} GOOD\_SIGNAL: \textit{Being in the main hall implies a high probability of strong signal strength due to the presence of a router}.
    \item \textbf{IF} GOOD\_SIGNAL \textbf{THEN} OFFLOADED: \textit{If the signal is strong, the robot proceeds with offloading its data}.
    \item \textbf{IF} OFFLOADED \textbf{THEN} BATTERY\_SAVED: \textit{Successful offloading reduces the robot's energy consumption, improving battery efficiency}.
\end{itemize}
\end{small} 

with the a confidence level of:

\begin{itemize}
    \item $\mu = 0.9$ (90\% expected success rate)
    \item $\sigma^2 = 0.05^2$ (variance capturing uncertainty)
\end{itemize}

\vspace{0.2cm}

\textbf{Situation 1: Expected situation: }\textit{The initial line of thought and corresponding beliefs are consistent with reality, as the robot successfully offloaded the corresponding task in the MAIN\_HALL.}
\begin{itemize}
    \item The robot detects 45 out of 50 objects by successfully offloading tasks and experiences good performance as expected.
    
    \item For \textbf{R3} only, the area encompassing the Main Hall keeps increasing its attention in both sigma and mu values using the beta update:
\begin{equation}
\alpha' = \alpha + S
\end{equation}
\begin{equation}
\beta' = \beta + F
\end{equation}

\textbf{Final Posterior:} Updated Mean: $\mu' \approx 0.90$, Updated Variance: $\sigma'^2 \approx 0.0019$. 
\end{itemize}
\vspace{0.2cm}

\textbf{Situation 2 - Unexpected Situation: }\textit{The callbox mini base station is moved to a different room (office 3), some of the objects are also moved away}

Under these circumstances, although the robot remains in the MAIN\_HALL, the detection performance drops significantly. This decline is largely due to the Edge/Cloud detector's limitations caused by poor radio signal quality in the room. Meanwhile, the on-board detector fails to recognize all classes due to being out of distribution, ultimately detecting only 20 out of 50 unique objects. A robust meta-reasoner should recognize this unexpected situation and prompt the robot to move away from the MAIN\_HALL to find more unique objects from other locations as specified by the task.

\vspace{0.2cm}

\textbf{- R1: Generic meta-reasoning}: The generic meta-reasoner does not understand the concept of radio signal quality. Without considering signal quality, it can switch from the poorly performing edge detector back to the onboard detector and continue exploring within the MAIN\_HALL. During the task execution, 9 human interventions were applied to assist the robot's detection, mainly to drag the robot out of the poorly performing area and detect objects located outside using the Edge detector. Of these, 5 interventions regarding new objects were recognized by the meta-reasoner, while the remaining interventions related to signal strength were not recognized, as the reasoner does not understand the context of signal strength. A total of 29 unique objects were recognized, leading to a robustness of 83\% and a success rate to 58\%. The total battery drop from the beginning to the end of the task execution is 43.5\%, which averages to 1.55\% per unique object.

\vspace{0.2cm}

\textbf{- R2: Customized meta-reasoner with radio network quality}:   In this case, the concept of radio signal strength is manually defined as part of VoC to boost the performance of the reasoner. The robot can move to areas with higher radio signal quality (outside the MAIN\_HALL), significantly improving the success rate achieved. A total of 35 unique objects are recognized. During the process, there are 4 human interventions, all of which are recognized by the meta-reasoner. This leads to a robustness of 96\% and an increased success rate of 70\%. The battery consumption is 1.11\% per unique object.
\vspace{0.2cm}
    
\textbf{- R3: Meta reasoning with attention}: The semantic attention map is an unsupervised model with a belief distribution revised based on a binomial function. When the MAIN\_HALL exhibits a low expected utility, it is marked as a low-attention area, prompting the robot to update its belief and proactively explore other areas. As a result, the robot moves away from the low-attention area and refocuses on nearby rooms with less difference in expected utility. Since the reasoning is based on attention rather than symbolic representations, grounding issues are eliminated. This allows the robot to dynamically reallocate attention to rooms near the relocated router without altering their semantic interpretation. Consequently, the robot moves to Office 3, which is later identified as a more relevant area. The robot now detects a high number of objects. A total of 36 unique objects are recognized. There is only 1 human intervention during the process, which is recognized by the meta-reasoner. This leads to a robustness of 99 and increases the success rate to 72\%. The battery consumption is 1.06\% per unique object.
\vspace{0.2cm}

    \begin{figure}[h]
    \centering
    \subfloat[Signal quality map]{%
        \includegraphics[width=0.45\linewidth]{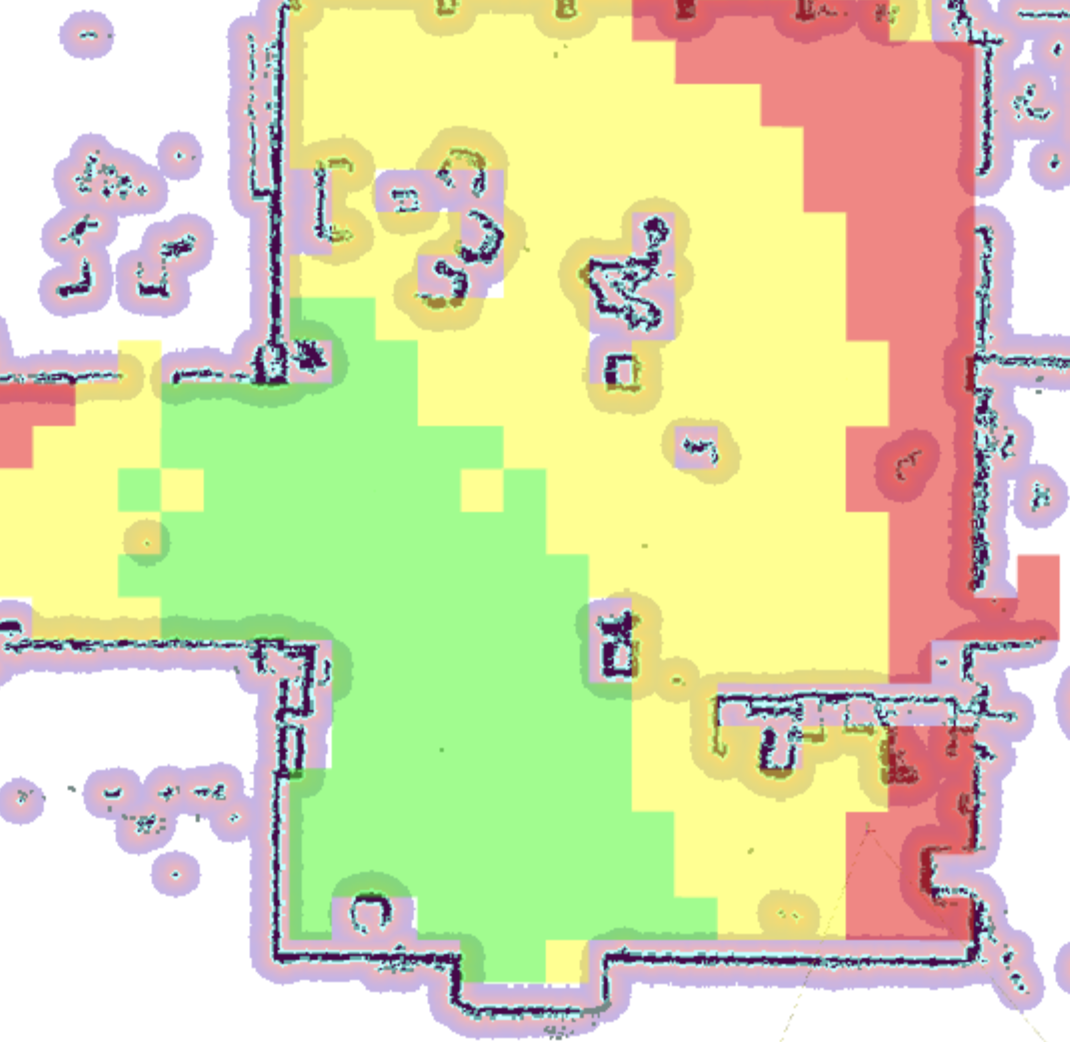}%
        \label{fig:case_study_1a}
    }
    \hfill
    \subfloat[Attention map before testbed position change]{%
        \includegraphics[width=0.45\linewidth]{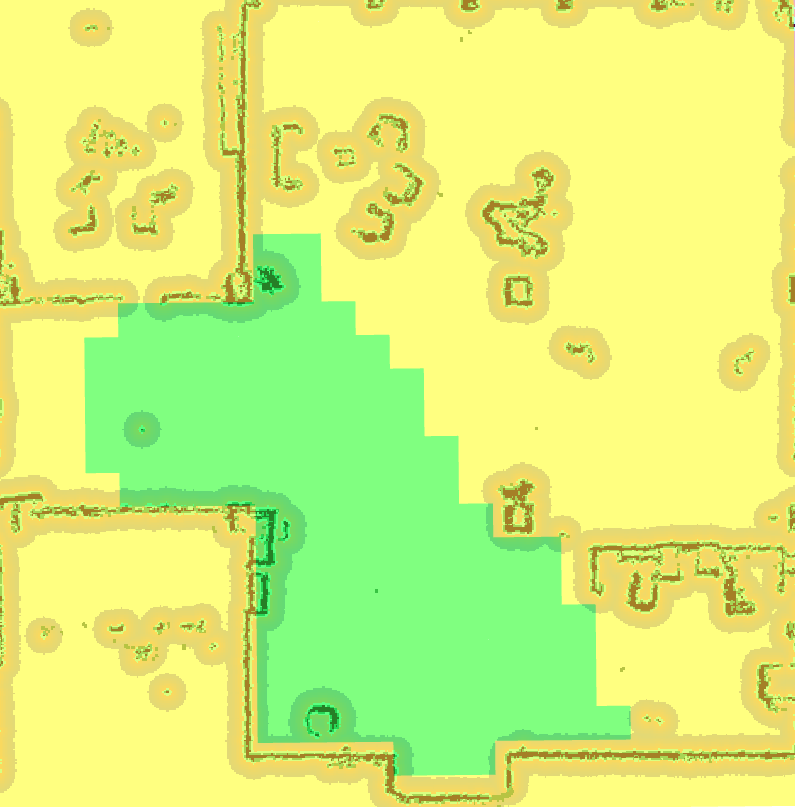}%
        \label{fig:case_study_1b}
    }
    \hfill
    \vspace*{2.0mm}
    \caption{Signal quality \& attention map for case study 1}
    \label{fig:case_study_1}
\end{figure}

\begin{table}[t]
\caption{Case Study 1 Results - \small \textbf{(R1)} Generic meta-reasoner, \textbf{(R2)} Customized meta-reasoner manually defined radio network quality and \textbf{(R3)} Meta reasoning with attention. Success Rate and Robustness: Higher is better. Battery Consumption: Lower is better.}
\label{tab:KPI1}
\centering
\label{tab:kpi_separate}
\resizebox{.8\columnwidth}{!}{%
\begin{tabular}{lccccc}
\toprule
\textbf{KPI}         & \textbf{R1} & \textbf{R2} & \textbf{ R3}  \\
\midrule
\textbf{KPI1}: Success Rate (\%)    & 58                      & 70                              & 72                                             \\
\textbf{KPI2}: Robustness           & 83                       & 96                              & 99                                             \\
\textbf{KPI3}: Battery Consumption (\%) & 1.55                & 1.11                             & 1.06                                         \\
\bottomrule
\end{tabular}%
}
\vspace{-3mm}
\end{table}

The radio signal quality map (Fig. \ref{fig:case_study_1a}) and the attention map (Fig. \ref{fig:case_study_1b}) illustrate the status of the reasoners R2 and R3 at the end of the operation. In the radio signal quality map, green indicates good quality, yellow represents average quality, and red indicates poor quality as defined in~\cite{5GERA_IROS}. In the attention map, green denotes high attention ($\mu$), encouraging cloud/edge operation offloading, while yellow indicates low attention, discouraging offloading. The confidence of the belief is stored per grid as ($\sigma$) but is not visualized on the map. In this case study, the focus area (green) of the radio signal quality map aligns with the attention (green) area of the attention map. It is self-evident that KPI1-3 listed in Table \ref{tab:KPI1} are consistent with the expected capabilities of the reasoner described in Table \ref{tab:comparison}. Therefore, both the proposed reasoner and the radio network quality customized reasoner work as expected under the unexpected situation of the case study.

\subsection{\textbf{Case Study 2: Mobile Robots Roaming Among Edges Using On-Demand Switching Over}} 

In this case study, the robot automatically switches computational resources between two networks: Edge 1 (located in the Main\_Hall, under a private 5G network enabled by Amari Callbox) and Edge 2 (located in the Office, under a local WiFi network). This switch is typically triggered at a meta-level based on the initial Line of Thought. The meta-policy behind this switching aims to enhance the success rate of robot cloud operations by ensuring the robot always utilizes the best available network in terms of quality. The details of the initial Line of Thought in Case Study 2 are outlined below:

\begin{footnotesize} 
\begin{itemize}
    \item \textbf{IF} MAIN\_HALL \textbf{THEN} CLOSE\_TO\_EDGE\_1  
    \item \textbf{IF} CLOSE\_TO\_EDGE\_1 \textbf{THEN} SWITCH\_TO\_EDGE\_1  
    \item \textbf{IF} SWITCH\_TO\_EDGE\_1 \textbf{THEN} REDUCED\_LATENCY  
    \item \textbf{IF} REDUCED\_LATENCY \textbf{THEN} IMPROVED\_DETECTION\_SUCCESS  
    \item \textbf{IF} OFFICE \textbf{THEN} CLOSE\_TO\_EDGE\_2  
    \item \textbf{IF} CLOSE\_TO\_EDGE\_2 \textbf{THEN} SWITCH\_TO\_EDGE\_2  
    \item \textbf{IF} SWITCH\_TO\_EDGE\_2 \textbf{THEN} REDUCED\_LATENCY  
    \item \textbf{IF} REDUCED\_LATENCY \textbf{THEN} IMPROVED\_DETECTION\_SUCCESS 
    \item \textbf{IF} BORDER THEN USE\_SIGNAL\_QUALITY\_FOR\_COMPARISON  
    
\end{itemize}
\end{footnotesize} 


\begin{figure}[h]
    \centering
    \subfloat[Attention map for Switching]{%
        \includegraphics[width=0.3\linewidth]{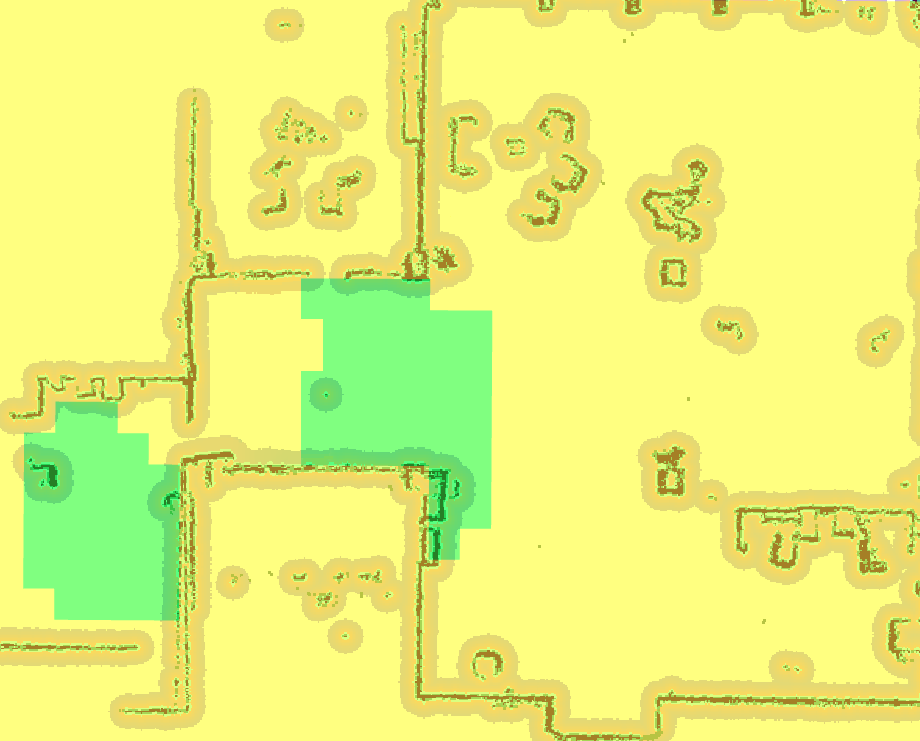}%
        \label{fig:case_study_2a}
    }
    \hfill
    \subfloat[SQM of WiFi near Office]{%
        \includegraphics[width=0.3\linewidth]{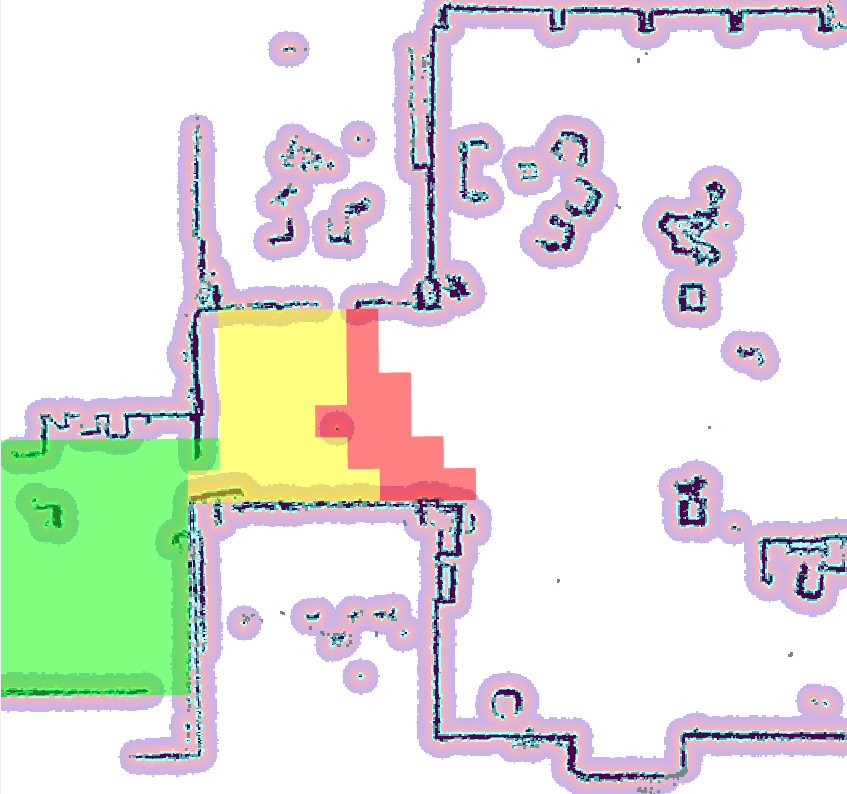}%
        \label{fig:case_study_2b}
    }
    \hfill
    \subfloat[SQM of 5G near Main hall]{%
        \includegraphics[width=0.3\linewidth]{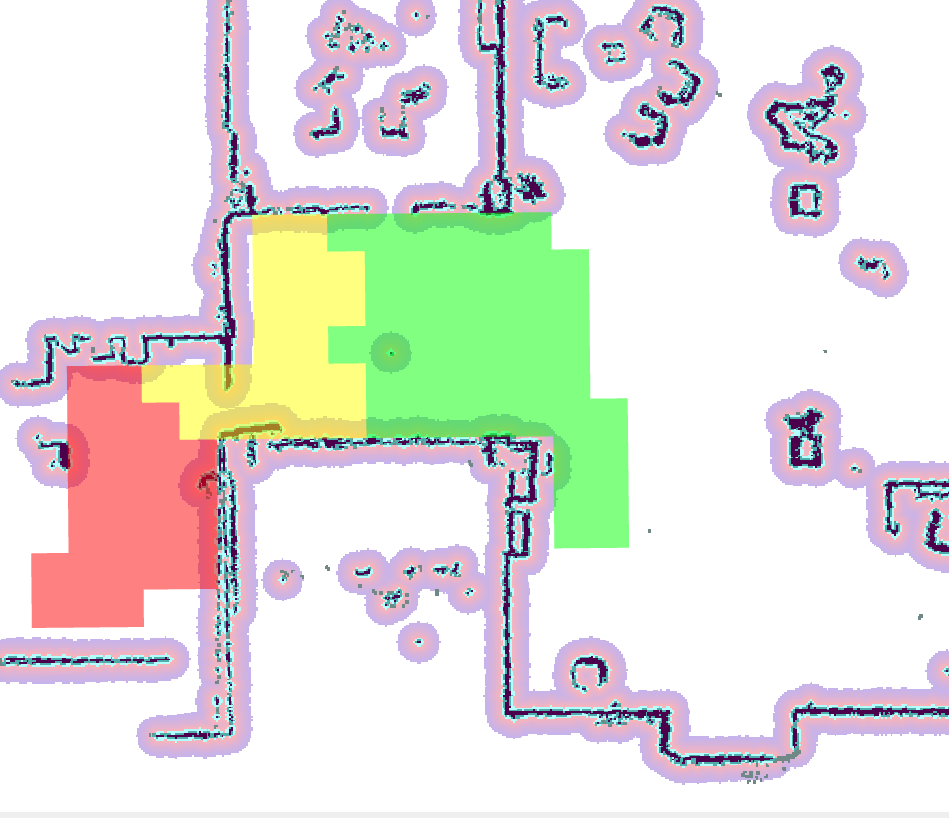}%
        \label{fig:case_study_2c}
    }
    \vspace{2.0mm}
    \caption{The attention map for Edge Switching and the Signal Quality Map (SQM) for WiFi near the office, as well as the private 5G network near the main hall.}
    
    \label{fig:case_study_2}
    
\end{figure}

\begin{table}[t]
\centering 
\caption{Case Study 2 Results - \small \textbf{(R1)} Generic meta-reasoner, \textbf{(R2)} Customized meta-reasoner manually defined radio network quality and \textbf{(R3)} Meta reasoning with attention. Availability: Higher is better.}
\label{tab:kpi_separate}
\resizebox{.7\columnwidth}{!}{%
\begin{tabular}{lccccc}
\toprule
\textbf{KPI}         & \textbf{R1} & \textbf{R2} & \textbf{R3}  \\
\midrule
\textbf{KPI4}: Availability (\%)     & 75        & 74       & 85               \\
\bottomrule
\end{tabular}%
}
\label{tab:KPI2}
\end{table}

Unexpected situations arise when the deployment time required by network applications exceeds expectations. If the robot repeatedly moves back and forth between Edge 1 and Edge 2 (for example, navigating near radio signal boundary areas as shown in the yellow areas in both Fig.\ref{fig:case_study_2b} and Fig.\ref{fig:case_study_2c}), the KPI4 Availability will decrease due to unnecessary redeployments. In this case, \textbf{KPI4} is determined by the time consumed by the switchover operations rather than the radio signal quality. As shown in Fig. \ref{fig:case_study_2a}, following the update, the attention near the border drops to yellow to discourage switching in that area, even if a difference in network signal quality is detected. Conversely, other areas of the rooms with edge devices exhibit high attention, promoting switching if a signal difference appears. The performance of the benchmarked reasoners is listed in Table \ref{tab:KPI2}, the proposed reasoner R3 recognizes these unexpected situations and generates control policies to reduce the amount of switchover in the marginal areas, subsequently improving availability. 

\subsection{\textbf{Discussion}}
The two case studies highlight the importance of managing network changes at the meta-level to enhance the robustness of cloud-based robot operations. Numerous studies, including~\cite{networkingstreaming,NetworkCloud,networkingROS} show that variations in radio network quality can significantly impact real-time object detection and decision-making. \textbf{Case Study 1} aligns with the results obtained in those studies, where both signal quality-aware (\textbf{R2}) and attention-based meta-reasoning (\textbf{R3)} outperform the generic meta-reasoner (\textbf{R1}) in unexpected situations. However, when additional uncertainty occurs at the object level, signal quality alone may not suffice for ensuring reliable cloud operations. As shown in \textbf{Case Study 2}, meta-reasoning with attention maps (\textbf{R3}) outperforms the signal quality-aware reasoner (\textbf{R2}) as well as the generic meta-reasoner (\textbf{R1}), demonstrating better scalability in unexpected situations. Overall, attention maps can be used to coordinate both immediate operational tasks and overarching strategic decisions, thereby enhancing the robot's adaptation to unexpected network changes, which is crucial for future connected robots.

\section{Conclusion}
\label{sec:Conclusion}
This paper presents an enhanced meta-reasoning framework that improves adaptability in unexpected situations by introducing a novel decoupling of object-reasoning and meta-reasoning into two loosely connected reasoning loops. This separation allows robots to focus on high-level abstract factors, enabling more flexible decision-making independent of specific environmental conditions.
A key advancement in this work is the implementation of an unsupervised attention updating mechanism. This approach dynamically adjusts the focus of the meta-level strategy based on the quality of experience, allowing for continuous adaptation without requiring predefined rules. By leveraging a ROS-compatible semantic map, the system effectively perceives environmental changes through the quality of service, enabling real-time reasoning in dynamic settings.
Finally, the proposed framework is validated in cloud robotics scenarios, demonstrating its effectiveness in managing unexpected situations. By incorporating meta-reasoning into cloud-based decision-making, the system efficiently orchestrates task switching between local and remote operations while optimizing resource utilization. These results highlight the potential of attention-based meta-reasoning as a robust approach for enhancing autonomy and adaptability in robotic systems.


\section*{Acknowledgment}
The research leading to these results has been supported by the EU's H$2020$ 5G ERA Project (grant no. $101016681$).


\bibliographystyle{IEEEtran}
\bibliography{biblio}

\begin{thebibliography}{10}
\providecommand{\url}[1]{#1}
\csname url@samestyle\endcsname
\providecommand{\newblock}{\relax}
\providecommand{\bibinfo}[2]{#2}
\providecommand{\BIBentrySTDinterwordspacing}{\spaceskip=0pt\relax}
\providecommand{\BIBentryALTinterwordstretchfactor}{4}
\providecommand{\BIBentryALTinterwordspacing}{\spaceskip=\fontdimen2\font plus
\BIBentryALTinterwordstretchfactor\fontdimen3\font minus \fontdimen4\font\relax}
\providecommand{\BIBforeignlanguage}[2]{{%
\expandafter\ifx\csname l@#1\endcsname\relax
\typeout{** WARNING: IEEEtran.bst: No hyphenation pattern has been}%
\typeout{** loaded for the language `#1'. Using the pattern for}%
\typeout{** the default language instead.}%
\else
\language=\csname l@#1\endcsname
\fi
#2}}
\providecommand{\BIBdecl}{\relax}
\BIBdecl

\bibitem{Thrun2005}
S.~Thrun, W.~Burgard, and D.~Fox, \emph{Probabilistic robotics}.\hskip 1em plus 0.5em minus 0.4em\relax Cambridge, Mass.: MIT Press, 2005.

\bibitem{Herrmann2023}
J.~W. Herrmann, \emph{Metareasoning for Robots Adapting in Dynamic and Uncertain Environments}, 1st~ed.\hskip 1em plus 0.5em minus 0.4em\relax Springer, 2023.

\bibitem{Stuart1990}
S.~Russell and E.~Wefald, ``Principles of metareasoning,'' \emph{Artificial Intelligence}, vol.~49, pp. 361--395, 1990.

\bibitem{Cox2013}
M.~T. Cox and A.~Raja, \emph{Metareasoning: Thinking about Thinking}.\hskip 1em plus 0.5em minus 0.4em\relax MIT Press, 2013.

\bibitem{Griffiths2019}
T.~L. Griffiths, F.~Callaway, M.~B. Chang, E.~Grant, P.~M. Krueger, and F.~Lieder, ``Doing more with less: meta-reasoning and meta-learning in humans and machines,'' \emph{Current Opinion in Behavioral Sciences}, vol.~29, pp. 24--30, 2019.

\bibitem{Conitzer2013}
V.~Conitzer, ``Metareasoning as a formal computational problem,'' in \emph{Metareasoning: Thinking about Thinking}.\hskip 1em plus 0.5em minus 0.4em\relax MIT Press, 2013.

\bibitem{Qiu2012}
R.~Qiu, A.~Noyvirt, Z.~Ji, A.~Soroka, D.~Li, B.~Liu, G.~Arbeiter, F.~Weisshardt, and S.~Xu, ``Integration of symbolic task planning into operations within an unstructured environment,'' \emph{International Journal of Intelligent Mechatronics and Robotics}, vol.~2, p.~38, 2012.

\bibitem{Liu2012}
B.~Liu, D.~Li, R.~Qiu, Y.~Yue, C.~Maple, and S.~Gu, ``Fuzzy optimisation based symbolic grounding for service robots,'' in \emph{IEEE International Conference on Intelligent Robots and Systems}, 2012.

\bibitem{Gurevitch2018}
J.~Gurevitch, J.~Koricheva, and S.~Nakagawa, ``Meta-analysis and the science of research synthesis,'' \emph{Nature}, vol. 555, p. 175–182, 2018.

\bibitem{Epstein2013}
S.~L. Epstein and S.~Petrovic, ``Learning expertise with bounded rationality and self-awareness,'' 2013.

\bibitem{Kumar2024}
S.~Kumar, A.~Sharma, and V.~e.~a. Shokeen, ``Meta-learning for real-world class incremental learning: a transformer-based approach,'' \emph{Scientific Reports}, vol.~14, p. 23092, 2024.

\bibitem{Parashar2018}
P.~Parashar, A.~K. Goel, B.~Sheneman, and H.~I. Christensen, ``Towards life-long adaptive agents: using metareasoning for combining knowledge-based planning with situated learning,'' \emph{The Knowledge Engineering Review}, 2018.

\bibitem{5GERA_IROS}
A.~Lendinez, L.~Zanzi, S.~Moreno, G.~Gari, X.~Li, R.~Qiu, and X.~Costa-Perez, ``{Enhancing 5G-enabled Robots Autonomy by Radio-Aware Semantic Maps},'' \emph{IEEE IROS}, Oct. 2023.

\bibitem{5GERA2024}
\BIBentryALTinterwordspacing
{5G-ERA Consortium}, ``{PPDR} healthcare netapp verification report,'' 5G-ERA Project, Tech. Rep. D6.2, 2024. [Online]. Available: \url{https://5g-era.eu/wp-content/uploads/2024/09/D6.2_PPDR-healthcare-NetApp-verification-report_v1.1.pdf}
\BIBentrySTDinterwordspacing

\bibitem{RAG2020}
P.~Lewis, E.~Perez, A.~Piktus, F.~Petroni, V.~Karpukhin, N.~Goyal, H.~Küttler, M.~Lewis, W.~tau Yih, T.~Rocktäschel, S.~Riedel, and D.~Kiela, ``Retrieval-augmented generation for knowledge-intensive nlp tasks,'' in \emph{NIPS'20: Proceedings of the 34th International Conference on Neural Information Processing Systems}, 2020, pp. 9459 -- 9474.

\bibitem{SUMMIT-XL}
\BIBentryALTinterwordspacing
Robotnik. (2025) {SUMMIT-XL Mobile Robot}. Accessed on Feb 2025. [Online]. Available: \url{https://robotnik.eu/products/mobile-robots/summit-xl-en-2/}
\BIBentrySTDinterwordspacing

\bibitem{Amarisoft}
\BIBentryALTinterwordspacing
AmariSoft. (2025) {AMARI Callbox Mini}. Accessed on Feb 2025. [Online]. Available: \url{https://www.amarisoft.com/test-and-measurement/device-testing/device-products/amari-callbox-mini}
\BIBentrySTDinterwordspacing

\bibitem{ROSWiki}
{ROS Wiki}, ``Connected robotics platform,'' Retrieved February 26, 2025, from \url{http://wiki.ros.org/Connected%20Robotics%20Platform}, 2024.

\bibitem{SemanticMapROSPacakge}
\BIBentryALTinterwordspacing
5G-ERA. (2025) {Signal Quality Network Application ROS2}. Accessed on Mar 2025. [Online]. Available: \url{https://github.com/5G-ERA/signalQualityNetworkApplicationRos2}
\BIBentrySTDinterwordspacing

\bibitem{networkingstreaming}
H.~Aagela, V.~Holmes, M.~Dhimish, and D.~Wilson, ``\BIBforeignlanguage{English}{Impact of video streaming quality on bandwidth in humanoid robot nao connected to the cloud},'' in \emph{\BIBforeignlanguage{English}{Proceedings of the Second International Conference on Internet of things, Data and Cloud Computing ; Conference date: 22-03-2017 Through 23-03-2017}}, Mar. 2017.

\bibitem{NetworkCloud}
S.~Chinchali, A.~Sharma, J.~Harrison, A.~Elhafsi, D.~Kang, E.~Pergament, E.~Cidon, S.~Katti, and M.~Pavone, ``Network offloading policies for cloud robotics: A learning‐based approach,'' \emph{Autonomous Robots}, no.~45, 2021.

\bibitem{networkingROS}
\BIBentryALTinterwordspacing
A.~Botta, J.~Cacace, R.~De~Vivo, B.~Siciliano, and G.~Ventre, ``Networking for cloud robotics: The dewros platform and its application,'' \emph{Journal of Sensor and Actuator Networks}, vol.~10, no.~2, 2021. [Online]. Available: \url{https://www.mdpi.com/2224-2708/10/2/34}
\BIBentrySTDinterwordspacing

\end{thebibliography}

\end{document}